\documentclass[11pt,a4paper]{article}
\usepackage[hyperref]{acl2019}
\usepackage{times}
\usepackage{latexsym}

\usepackage{url}

\usepackage{times}
\usepackage{latexsym}
\usepackage{url}
\usepackage{float}

\usepackage{amsmath,graphicx,amsfonts,amssymb,algorithm,algorithmicx}
\usepackage{arydshln}
\usepackage{siunitx} 
\usepackage{booktabs}
\usepackage{microtype}

\usepackage{graphicx}
\usepackage{subcaption}
\usepackage{caption}
\usepackage{lipsum}
\usepackage{placeins}
\aclfinalcopy % Uncomment thiss line for the final submission
\def\aclpaperid{88} %  Enter the acl Paper ID here

\title{Modality-based Factorization for Multimodal Fusion}

\author{Elham J. Barezi, Pascale Fung \\
Center for Artificial Intelligence Research (CAiRE)\\
Department of Computer Science and  Engineering \\
The Hong Kong University of Science and Technology,  Clear Water Bay, Hong Kong \\
  \texttt{ejs@cse.ust.hk,pascale@ust.hk} }

\date{}

\begin{document}
\maketitle
\begin{abstract}
We propose a novel method, Modality-based Redundancy Reduction Fusion (MRRF), for understanding and modulating the relative contribution of each modality in multimodal inference tasks. 
This is achieved by obtaining an $(M+1)$-way tensor to consider the high-order relationships between $M$ modalities and the output layer of a neural network model. 
Applying a modality-based tensor factorization method, which adopts different factors for different modalities, results in removing  information present in a modality that can be compensated by other modalities, with respect to model outputs.
This helps to understand the relative utility of information in each modality.   
In addition it leads to a less complicated model with less parameters and therefore could be applied as a regularizer avoiding overfitting. 
We have applied this method to three different multimodal datasets in sentiment analysis, personality trait recognition, and emotion recognition. 
We are able to recognize relationships and relative importance of different modalities in these tasks and 
achieves a 1\% to 4\% improvement on several evaluation measures compared to the state-of-the-art for all three tasks. 
\end{abstract}

\section{Introduction}\label{sec:intro}

Multimodal data fusion is a desirable method for many machine learning tasks
where information is available from multiple source modalities, 
typically achieving better predictions through integration of information from different modalities. 
Multimodal integration can handle missing data from one or more modalities. 
Since some modalities can include noise, 
it can also lead to more robust prediction. 
Moreover, since some information may not be visible in some modalities or a single modality may not be powerful enough for a specific task, considering multiple modalities often improves performance~\cite{potamianos2003recent,soleymani2012multimodal,kampman2018investigating}.  

For example, humans assign personality traits to each other, as well as to virtual characters by inferring personality from diverse cues, both behavioral and verbal, 
suggesting that a model to predict personality should take into account multiple modalities such as language, speech, and visual cues. 

Our method, Modality-based Redundancy Reduction multimodal Fusion (MRRF),  
builds on recent work in mutimodal fusion utilizing first an outer product tensor of input modalities to better capture inter-modality dependencies~\cite{zadeh2017tensor} and a recent approach to reduce the number of elements in the resulting tensor through low rank  factorization~\cite{liu2018efficient}. 
Whereas the factorization used in~\cite{liu2018efficient} utilizes a single compression rate across all modalities, we instead use Tuckers tensor decomposition (see the Methodology section), which allows different compression rates for each modality. 
This allows the model to adapt to variations in the amount of useful information between modalities. 
Modality-specific factors are chosen by maximizing performance on a validation set. 

% We propose a multimodal data fusion method which applys a $M+1$-way weight tensor $W$ to the $M$ dimentional modality integration tensor to consider the high-order relationship between $M$ input modalities. 

Applying a modality-based factorization method results in removing redundant information  duplicated across modalities and 
leading to fewer parameters with minimal information loss. 
Through maximizing performance on a validation set, 
our method can work as a regularizer, leading to a less complicated model and reducing overfitting. 
In addition, our modality-based factorization approach helps to understand the 
differences in useful information between modalities 
for the task at hand. 

We evaluate the performance of our approach using sentiment analysis, personality detection, and emotion recognition from audio, text and video frames. The method reduces the number of parameters which requires fewer training samples, providing efficient training for the smaller datasets, and accelerating both training and prediction. 
Our experimental results demonstrate that the proposed approach can make notable improvements, in terms of accuracy, mean average error (MAE), correlation, and $\text{F}_1$ score, especially for the applications with more complicated inter-modality relations.   

We further study the effect of different compression rates for different modalities. Our results on the importance of each modality for each task supports the previous results on the usefulness of each modality for personality recognition, emotion recognition and sentiment analysis. 
% Moreover, our results demonstrate that our factorization approach avoids underfitting and overfitting for very simple and very large models, respectively.  

In the sequel, we first describe related work. 
% we first explain the notation used in this paper. 
We elaborate on the details of our proposed method in Methodology section. 
In the following section we go on to describe our experimental setup. 
In the Results section, we compare the performance of MRRF and state-of-the-art baselines on three different datasets and discuss the effect of compression rate on each modality. 
Finally, we provide a brief conclusion of the approach and the results. 
Supplementary materials describe the methodology in greater detail. 

\paragraph{Notation} \label{sec:notation} 
The operator $\otimes$ is the outer product operator where $z_1 \otimes \ldots \otimes z_M$ for $z_i \in \mathbb{R}^{d_i}$ leads to a M-way tensor in $\mathbb{R}^{d_1 \times \ldots\times d_M}$.
The operator $\times_k$, for a given $k$, is k-mode product of a tensor $R \in \mathbb{R}^{r_{1} \times r_{2} \times \ldots \times r_{M}}$ and a matrix $W \in \mathbb{R}^ {d_{k} \times r_{k}}$ as $W \times_k R$, which results in a tensor $\bar{R} \in \mathbb{R}^{r_{1} \times \ldots \times r_{k-1} \times d_{k} \times r_{k+1} \times \ldots \times r_{M}}$. 
% This operator first flattens the tensor $R$ and converts it to a matrix $\hat{R} \in R^{r_k \times r_{1}r_2 \ldots r_{k-1} r_{k+1} \ldots r_{M}}$. The next step is a simple matrix product as $W \hat{R} \in \mathbb{R}^{d_k \times r_{1} \ldots r_{k-1} r_{k+1} \ldots r_{M}}$. By unflattening the resulted matrix, we can convert it to a tensor in $\mathbb{R}^{r_{1} \times \ldots \times r_{k-1} \times d_{k} \times r_{k+1} \times \ldots \times r_{M}}$.

\section{Related Work}
\paragraph*{Multimodal Fusion:}
Multimodal fusion~\cite{Ngiam2011Multimodal} has a very broad range of applications, including audio-visual speech recognition 
\cite{potamianos2003recent}, classification of images and their captions~\cite{Srivastava2012Multimodal}, multimodal emotion recognition \cite{soleymani2012multimodal}, medical image analysis \cite{james2014medical}, multimedia event detection \cite{lan2014multimedia}, personality trait detection \cite{kampman2018investigating}, and sentiment analysis \cite{zadeh2017tensor}.

According to the recent work by \cite{baltruvsaitis2018multimodal}, the techniques for multimodal fusion can be divided into early, late and hybrid approaches. 
Early approaches combine the multimodal features immediately by simply concatenating them~\cite{d2015review}. Late fusion combines the decision for each modality (either classification, or regression), by voting \cite{morvant2014majority}, averaging \cite{shutova2016black} or weighted sum of the outputs of the learned models \cite{glodek2011multiple,shutova2016black}. The hybrid approach combines the prediction by early fusion and unimodal predictions.

It has been observed that early fusion (feature level fusion) concentrates on the inter-modality information rather than intra-modality information~\cite{zadeh2017tensor} due to the fact that inter-modality information can be more complicated at the feature level and dominates the learning process. 
On the other hand, these fusion approaches are not powerful enough to extract the inter-modality integration model and they are limited to some simple combining methods~\cite{zadeh2017tensor}. 

\newcite{zadeh2017tensor} proposed combining $n$ modalities by computing an $n$-way tensor as a tensor product of the $n$ different modality representations followed by a flattening operation, in order to include 1-st order to n-th order inter modality relations. 
This is then fed to a neural network model to make predictions.
The authors show that their proposed method improves the accuracy by considering both inter-modality and intra-modality relations. However, the generated representation has a very large dimension which leads to a very large hidden layer and therefore a huge number of parameters. 

The authors of \cite{poria2017context, poria2017multi, zadeh2018memory, zadeh2018multi} introduce attention mechanisms utilizing the contextual information available from the utterances for each speaker. They require additional information like the identity of the speaker, the sequence of the utterance-sentiments while integrating the multimodal data. Since these methods, despite our proposed method, need additional information might not be available in the general scenario, we do not include them in our experiments.

\paragraph{Low Rank Factorization:}
Recently \cite{liu2018efficient} proposed a factorization approach in order to achieve a factorized version of the weight matrix which leads to fewer parameters while maintaining model accuracy. 
They use a CANDECOMP/PARAFAC decomposition~\cite{carroll1970analysis,harshman1970foundations} which follows Eq. \ref{eq:cp} in order to decompose a tensor $W \in \mathbb{R}^{d_{1}\times \ldots d_{M}}$ to several 1-dimensional vectors $w_m^i \in \mathbb{R}^{d_{k}}$:
\begin{equation}\label{eq:cp}
\begin{aligned} 
   W  &= \sum_{i=1}^r \lambda_i w_1^i \otimes w_2^i \otimes \ldots \otimes w_M^i \\
      &= \sum_{i=1}^r  \lambda_i \otimes_{m=1}^M w_m^{i}
\end{aligned}
\end{equation}
where $\otimes$ is the outer product operator, $\lambda_i$s are scalar weights to combine rank 1 decompositions.  
This approach used the same compression rate  for all modalities, i.e. $r$ is shared for all the modalities, and is not able to allow for varying compression rates between modalities. 
Previous studies have found that some modalities are more informative than others~\cite{de1997facial,kampman2018investigating}, suggesting that allowing different compression rates for different modalities should improve performance.

\section{Methodology} \label{sec:methodology}
\subsection{Tucker Factorization for Multimodal Learning}
\paragraph{Modality-based Redundancy Reduc- tion Fusion (MRRF):}
We have used Tucker's tensor decomposition method \cite{tucker1966some,hitchcock1927expression} which decomposes an $M$-way tensor $W \in \mathbb{R}^{d_{1} \times d_{2} \times \ldots \times d_{M}}$ 
to a core tensor $R \in \mathbb{R}^{r_{1} \times r_{2} \times \ldots \times r_{M}}$ and $M$ matrices $W_i \in \mathbb{R}^{r_{i} \times d_{i}}$, with $r_i\leq d_i$, as it can be seen in Eq. \ref{eq:tucker}. 
\begin{equation}\label{eq:tucker}
\begin{aligned}
  W  & = R \times_1 W_1 \times_2 W_2 \times_3 \ldots \times_M W_M, \\
       & W \in \mathbb{R}^{d_{1} \times d_{2} \times \ldots \times d_{M}} \\
       & R \in \mathbb{R}^{r_{1} \times r_{2} \times \ldots \times r_{M}}, \\
      & W_i \in  \mathbb{R}^ {d_{i} \times r_{i}}
\end{aligned}
\end{equation}
The operator $\times_k$ is a k-mode product of a tensor $R \in \mathbb{R}^{r_{1} \times r_{2} \times \ldots \times r_{M}}$ and a matrix $W \in \mathbb{R}^ {d_{k} \times r_{k}}$ as $R \times_k W_k$, which results in a tensor $\bar{R} \in \mathbb{R}^{r_{1} \times \ldots \times r_{k-1} \times d_{k} \times r_{k+1} \times \ldots \times r_{M}}$. 

For $M$ modalities with representations $D_1$, $D_2$, $\ldots$ and $D_M$ of size $d_1$, $d_2$, $\ldots$ and $d_M$, an $M$-modal tensor fusion approach as proposed by the authors of \cite{zadeh2017tensor} leads to a tensor $D = D_1 \otimes D_2 \otimes \ldots \otimes D_m \in \mathbb{R}^{d_1 \times d_2 \times \ldots \times d_M}$. 
The authors proposed flattening the tensor layer in the deep network which results in loss of the information included in the tensor structure. 
In this paper, we propose to avoid the flattening and 
follow Eq.~\ref{eq:dens} with weight tensor $W \in \mathbb{R}^{h \times d_1 \times d_2 \times \ldots \times d_M}$, where leads to an output layer $H$ of size $h$.
\begin{equation}\label{eq:dens}
H =  W D
\end{equation}

The above equation suffers from a large number of parameters ($O(\prod_{i=1}d_i h)$) which requires a large number of the training samples, huge time and space, and easily overfits. In order to reduce the number of parameters, we propose to use Tucker's tensor decomposition  \cite{tucker1966some,hitchcock1927expression} as shown in Eq. \ref{eq:decomp1}, which works as a low-rank regularizer \cite{fazel2002matrix}.

\begin{equation}\label{eq:decomp1}
% \times_M W_M 
\begin{aligned} 
    W &=  R \times_1 W_1 \times_2 W_2 \times_3 \ldots \times_{M+1} W_{M+1}, \\
      & W \in \mathbb{R}^{h \times d_1 \times d_2 \times \ldots \times d_M},   \\
      & R \in \mathbb{R}^{r_1 \times r_2 \times r_3 \times \ldots \times r_M}, \\
      & W_i \in \mathbb{R}^{r_i \times d_i},\, i=\lbrace 1,\ldots,M \rbrace , \\
      & W_{M+1} \in \mathbb{R}^{r_{M+1} \times h}
\end{aligned} 
\end{equation}

The non-diagonal core tensor $R$ maintain inter-modality information after compression, despite the factorization proposed by ~\cite{liu2018efficient} which loses part of inter-modality information.

% After combining the core tensor $R$ and the output matrix $W_{M+1}$, the decomposition in Eq. \ref{eq:decomp1} reduces to Eq. \ref{eq:decomp2}:

% \begin{equation}\label{eq:decomp2}
% \begin{aligned} 
%     W &=  \hat{R} \times_1 W_1 \times_2 \ldots \times_M W_M, \\
%       & \hat{R} \in \mathbb{R}^{r_1 \times ... \times r_M \times h} \\
% \end{aligned} 
% \end{equation}

% Substituting Eq.~\ref{eq:decomp2} into Eq.~\ref{eq:dens} leads to a factorized multimodal integration model. 
% Figure~\ref{fig:bimodal_ex} presents an example of this process with two input modalities. 

% \begin{figure}[!htb]
%       \centering
%       %\hspace*{-3.5cm} 
%       \includegraphics[trim=5cm 4.5cm 14.35cm 1.5cm,clip, scale=0.46]{img/mid_model.pdf}
%     %\vspace{-10mm}
%       \caption{\label{fig:bimodal_ex}The proposed tensor factorization layer for integrating two modalities.}
% \end{figure}
\subsection{Proposed MRRF framework}
We propose Modality-based Redundancy Reduction Fusion (MRRF), a tensor fusion and factorization method allowing for modality specific compression rates, combining the power of tensor fusion methods with a reduced parameter complexity. 
Without loss of generality, we will consider the number of modalities to be 3 in this discussion. 

Our method first forms an outer product tensor from input modalities $D$, then projects this via a tensor $W$ to a feature vector $H$ passed as input to a neural network which performs the desired inference task. 
\begin{equation}\label{eq:dens1}
H =  W D 
\end{equation}

The trainable projection tensor $W$ represents a large number of parameters, and in order to reduce this number, we propose to use Tucker's tensor decomposition  \cite{tucker1966some,hitchcock1927expression}, 
% as shown in Eq. \ref{eq:decomp1}, 
which works as a low-rank regularizer \cite{fazel2002matrix}. 
This results in a decomposition of $W$ into a core tensor $R$ of reduced dimensionality and three modality specific matrices $W_i$. 
\begin{equation}\label{eq:dens2}
W = R \times_1 W_1 \times_2 W_2 \times_3 W_3
\end{equation}
where $\times_k$ is a k-mode product of a tensor and a matrix. 
Equation~\ref{eq:dens1} can then be re-written 
\begin{align}\label{eq:dens3}
Z &=  W_1 \times_1 W_2 \times_2 W_3 \times_3 D \nonumber\\
H &=  Z R
\end{align}

\begin{figure*}[!htb]
      \centering
      \hspace*{-0.5cm} 
      \includegraphics[trim=0cm 4.0cm 0cm 0cm,clip, scale=0.45]{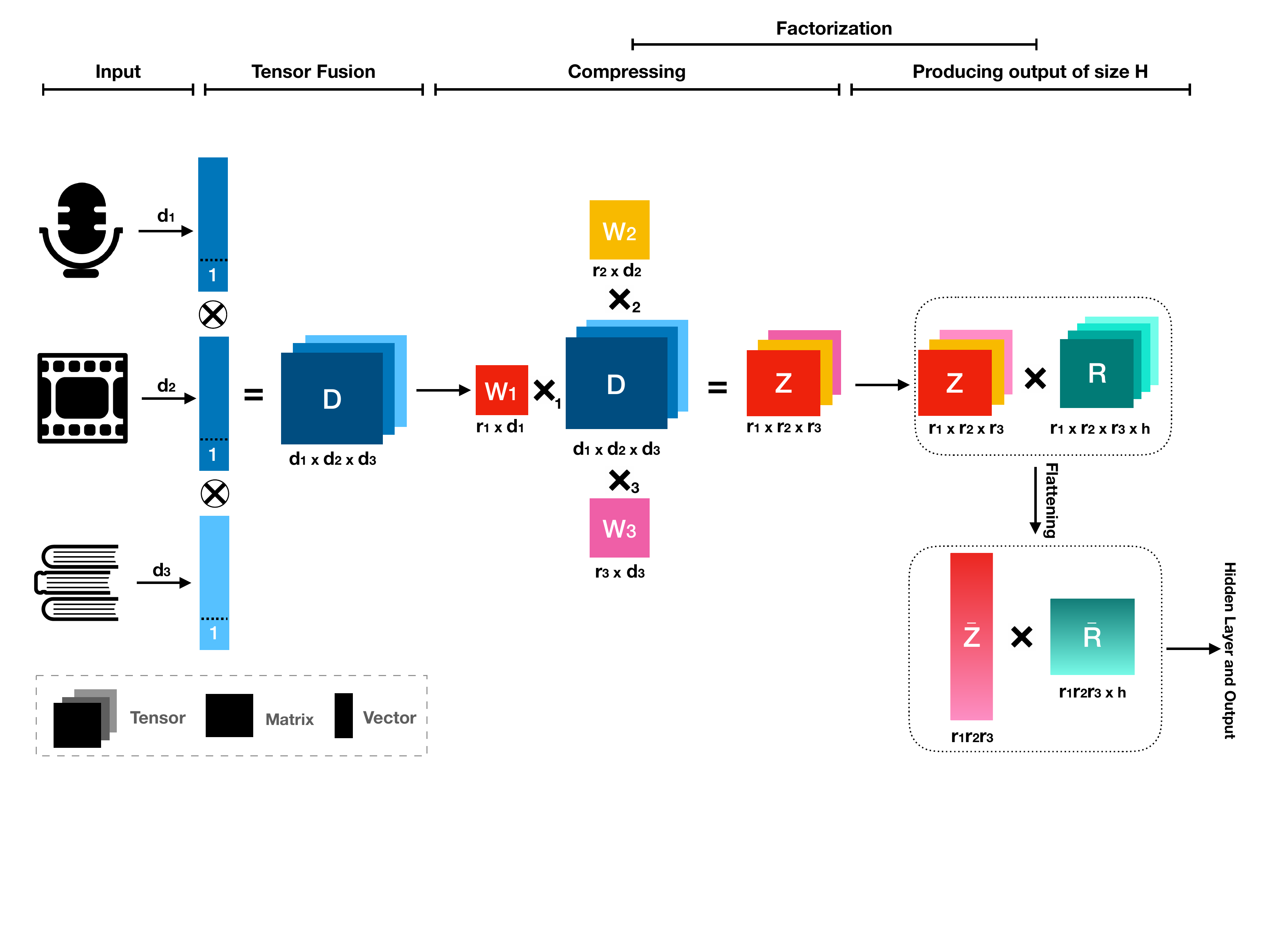}
    \vspace{-10mm}
      \caption{\label{fig:trimodal-architecture}Diagram of Modality-based Redundancy Reduction Multimodal Fusion (MRRF).}
\end{figure*}

See Figure~\ref{fig:trimodal-architecture} for an overview of this process for the case of three separate channels for audio, text, and video. 
In practice we flatten tensors $Z$ and $R$ to reduce this last operation to a matrix multiplication. 
Further details of the decomposition strategy can be found in the supplementary materials. 

Note that a simple outer product of the input features leads only to the high-order trimodal dependencies. 
In order to also obtain the unimodal and bimodal dependencies, the input feature vectors for each modality are padded by 1. 
This also provides a constant element whose corresponding factors in $W$ act as a bias vector. 

Algorithm~\ref{alg:1} shows the whole MRRF process.

%\vspace{-4mm}

\begin{algorithm}
\caption{Tensor Factorization Layer.}
\textbf{Input}: $n$ input modalities $D_1,\,D_2,\, \ldots ,\, D_n$ of size $d_1,\,d_2,\, \ldots ,\, d_n$, correspondingly. \\
\textbf{Initialization}: factorization size for each modality $r_1,\,r_2,\, \ldots ,\, r_n$.\\
\begin{algorithmic}[1]
\State Compute tensor $D=D_1\otimes D_2 \otimes \ldots \otimes D_n$
\State Generate the layers for $out=WD$ which $W =  \hat{R} \times_1 W_1 \times_2 \ldots \times_M W_M$ in order to transform the high-dimensional tensor $D$ to the output $h$.
\State Use Adam optimizer for the differentiable tensor factorization layer to find the unknown parameters $W_1,\,W_2,\, \ldots ,\, W_n,\, \hat{R}$.
\end{algorithmic}                        
\textbf{Output}: Factors for Weight Matrix $W$: $W_1,\,W_2,\, \ldots ,\, W_n,\, R$.
\label{alg:1}
\end{algorithm}

The original tensor fusion approach as proposed in \cite{zadeh2017tensor} flattened the tensor $D$ which results in loss of the information included in the tensor structure, which is avoided in our approach. 
\newcite{liu2018efficient} developed a similar approach to ours using a diagonal core tensor $R$, losing much inter-modality information. 
Our non-diagonal core tensor maintains key inter-modality information after compression. 

Note that the factorization step is task dependent, included in the deep network structure and learned during network training. Thus, for follow-up learning tasks, we would learn a new factorization specific to the task at hand, typically also estimating optimal compression ratios as described in the discussion section. In this process, any shared, helpful information is retained, as demonstrated by our results.  
 
\paragraph{Analysis of parameter complexity:}
Following our proposed approach, we have decomposed the trainable $W$ tensor to four substantially smaller trainable matrices ($W_1, \, W_2, \, W_3, \, R$) 
% For the feature level information of size $d_1$, $d_2$ and $d_3$ for three different modalities, this leads 
leading to $O(\sum_{i=1}^M(d_i*r_i)+\prod_{i=1}^Mr_i*h)$ parameters.
Concat fusion (CF) leads to a layer size of $O(\sum_{i=1}^Md_i)$ and $O(\sum_{i=1}^Md_i*h)$ parameters.

The tensor fusion approach (TF), 
% applying the flattening method directly to Eq. \ref{eq:dens}, 
leads to a layer size of $O(\prod_{i=1}^Md_i)$, and $O(\prod_{i=1}^Md_i*h)$ parameters. 
The LMF approach \cite{liu2018efficient} requires training $O(\sum_{i=1}^M r*h*d_i)$ parameters, where $r$ is the rank used for all the modalities. 
 
It can be seen that the number of parameters in the proposed approach is substantially fewer than the simple tensor fusion (TF) approach and comparable to the LMF approach.

\section{Experimental Setup} \label{sec:experiments_setup}
\subsection{Datasets}

We perform our experiments on the following multimodal datasets: CMU-MOSI \cite{zadeh2016mosi}, POM \cite{park2014computational}, and IEMOCAP \cite{busso2008iemocap} for sentiment analysis, speaker traits recognition, and emotion recognition, respectively. 
These tasks can be done by integrating both verbal and nonverbal behaviors of the persons.

The CMU-MOSI dataset is annotated on a seven-step scale as highly negative, negative, weakly negative, neutral, weakly positive, positive, highly positive which can be considered as a 7 class classification problem with 7 labels in the range $[-3,+3]$. The dataset is an annotated dataset of 2199 opinion utterances from 93 distinct YouTube movie reviews, each containing several opinion segments.  Segments average of 4.2 seconds in length.

The POM dataset is composed of 903 movie review videos. Each video is annotated with the following speaker traits: confident, passionate, voice pleasant, dominant, credible, vivid, expertise, entertaining, reserved, trusting, relaxed, outgoing, thorough, nervous, persuasive and humorous. 

The IEMOCAP dataset is a collection of 151 videos of recorded dialogues, with 2 speakers per session for a total of 302 videos across the dataset. Each segment is annotated for the presence of 9 emotions (angry, excited, fear, sad, surprised, frustrated, happy, disgust and neutral). 

Each dataset consists of three modalities, namely language, visual, and acoustic.
The visual and acoustic features are calculated by taking the average of their feature values over the word time interval \cite{chen2017multimodal}. In order to perform time alignment across modalities, the three modalities are aligned using P2FA \cite{yuan2008speaker} at the word level.

Pre-trained 300-dimensional Glove word embeddings \cite{chen2017multimodal} were used to extract the language feature representations, which encodes a sequence of the transcribed words into a sequence of vectors.

Visual features for each frame (sampled at 30Hz) are extracted using the library Facet\footnote{goo.gl/1rh1JN} which includes 20 facial action units, 68 facial landmarks, head pose, gaze tracking and HOG features \cite{zhu2006fast}.

COVAREP acoustic analysis framework \cite{degottex2014covarep} is used to extract low-level acoustic features, including 12 Mel frequency cepstral coefficients (MFCCs), pitch, voiced/unvoiced segmentation, glottal source, peak slope, and maxima dispersion quotient features. 

To evaluate model generalization, all datasets are split into training, validation, and test sets such that the splits are speaker independent, i.e., no speakers from the training set are present in the test sets. Table \ref{tab:data} illustrates the data splits for all the datasets in detail.

\begin{table}[!htb]
\centering
\small
\scalebox{0.8}{
\begin{tabular}{c|c|c|c}
\toprule
Dataset & CMU-MOSI & IEMOCAP & POM \\
Level & Segment & Segment & Video \\
\midrule
Train & 1284 & 6373 &  600 \\
Valid & 229  & 1775 &  100  \\
Test & 686   & 1807 &  203   \\
\bottomrule
\end{tabular}
}
\caption{\label{tab:data} The speaker independent data splits for training, validation, and test sets}
\end{table}

% NOTE: table moved from Results section so it appears at top of page with Results section heading
\begin{table*}[htb]
%\vspace{-1cm}
\centering
\small
\scalebox{0.8}{
\begin{tabular}{c|ccccc|ccc|c|c|c|c}
    \toprule
    Dataset &  \multicolumn{5}{c|}{CMU-MOSI} & \multicolumn{3}{c}{POM}& \multicolumn{4}{|c}{IEMOCAP}\\
    Metric &MAE &Corr &Acc-2 &F1 &Acc-7 &MAE &Corr &Acc &F1-Happy &F1-Sad &F1-Angry &F1-Neutral \\
    \midrule 
    CF  &1.140 &0.52 &72.3 &72.1 &26.5 &0.865 &0.142 &34.1 &81.1 &81.2 &65.1 &44.1 \\
    TFN &0.970 &0.633 &73.9 &73.4 &32.1 &0.886 &0.093 &31.6 &83.6 &82.8 &84.2 &65.4 \\
    LMF &0.912 &0.668 &76.4 &75.7 &32.8 &0.796 &0.396 &42.8 &85.8 &85.9 &89.0 &71.7 \\
    MRRF &0.912 &0.772 &77.46 &76.73 &33.02 &0.69 &0.44 &43.02 &87.71 &85.9 &90.02 &73.7 \\
    \bottomrule

\end{tabular}
}
\caption{\label{tab:sent_table} Results for Sentiment Analysis on CMU-MOSI, emotion recognition on IEMOCAP and personality trait recognition on POM. (CF, TF, and LMF stand for concat, tensor and low-rank fusion respectively). }
\end{table*}

\subsection{Model Architecture}
% As it has been done by 
Similarly 
to~\cite{liu2018efficient}, we use a simple model architecture for extracting the representations for each 
% of the single modalities. 
modality. 
% For three modalities including audio, visual and language, 
We used three unimodal sub-embedding networks to extract representations $z_a$, $z_v$ and $z_l$ for each modality, respectively. For acoustic and visual modalities, the sub-embedding network is a simple 2-layer feed-forward neural network, and for language, we used a long short-term memory (LSTM) network \cite{hochreiter1997long}. 

% We have tuned each of the parameters, including
We tuned 
the layer sizes, the learning rates and the compression rates, by checking the mean average error for the validation set by grid search. 
We trained our model using the Adam optimizer \cite{kingma2014adam}. All models were implemented with Pytorch \cite{paszke2017automatic}.

\section{Experimental Results and Comparing with State-of-the-art }\label{sec:experiments_results}
We compared our proposed method with three baseline methods. 
Concat fusion (CF) \cite{baltruvsaitis2018multimodal} proposes a simple concatenation of the different modalities followed by a linear combination.  
The tensor fusion approach (TF) \cite{zadeh2017tensor} computes a tensor including uni-modal, bi-modal, and tri-modal combination information. 
LMF~ \cite{liu2018efficient} is a tensor fusion method that performs tensor factorization using the same rank for all the modalities in order to reduce the number of parameters. Our proposed method aims to use different factors for each modality.

In Table~\ref{tab:sent_table}, we present mean average error (MAE), the correlation between prediction and true scores, binary accuracy (Acc-2), multi-class accuracy (Acc-7) and F1 measure. 
The proposed approach outperforms baseline approaches in nearly all metrics, with marked improvements in Happy and Neutral recognition. The reason is that the inter-modality information for these emotions is more complicated than the other emotions and requires a non-diagonal core tensor to extract the complicated information. 
It is worth to note that for the equivalent setting and equal ranks for all the modalities, the result of the proposed method is always marginally better than LMF method.

\subsection{Investigating the Effect of Compression Rate on Each Modality}\label{sec:experiments_rate}
In this section, we aim to investigate the amount of redundant information in each modality. To do this, after obtaining a tensor which includes the combinations of all modalities with the equivalent size, we factorize a single dimension of the tensor while keeping the size for the other modalities fixed. By observing how the performance changes by compression rate, one can find how much redundant information is contained in the corresponding modality relative to the other modalities. 

\begin{figure}[!htb]

      \centering
      \hspace*{-.5cm} 
      \includegraphics[trim=2cm 3cm 10cm 3cm,clip, scale=0.4]{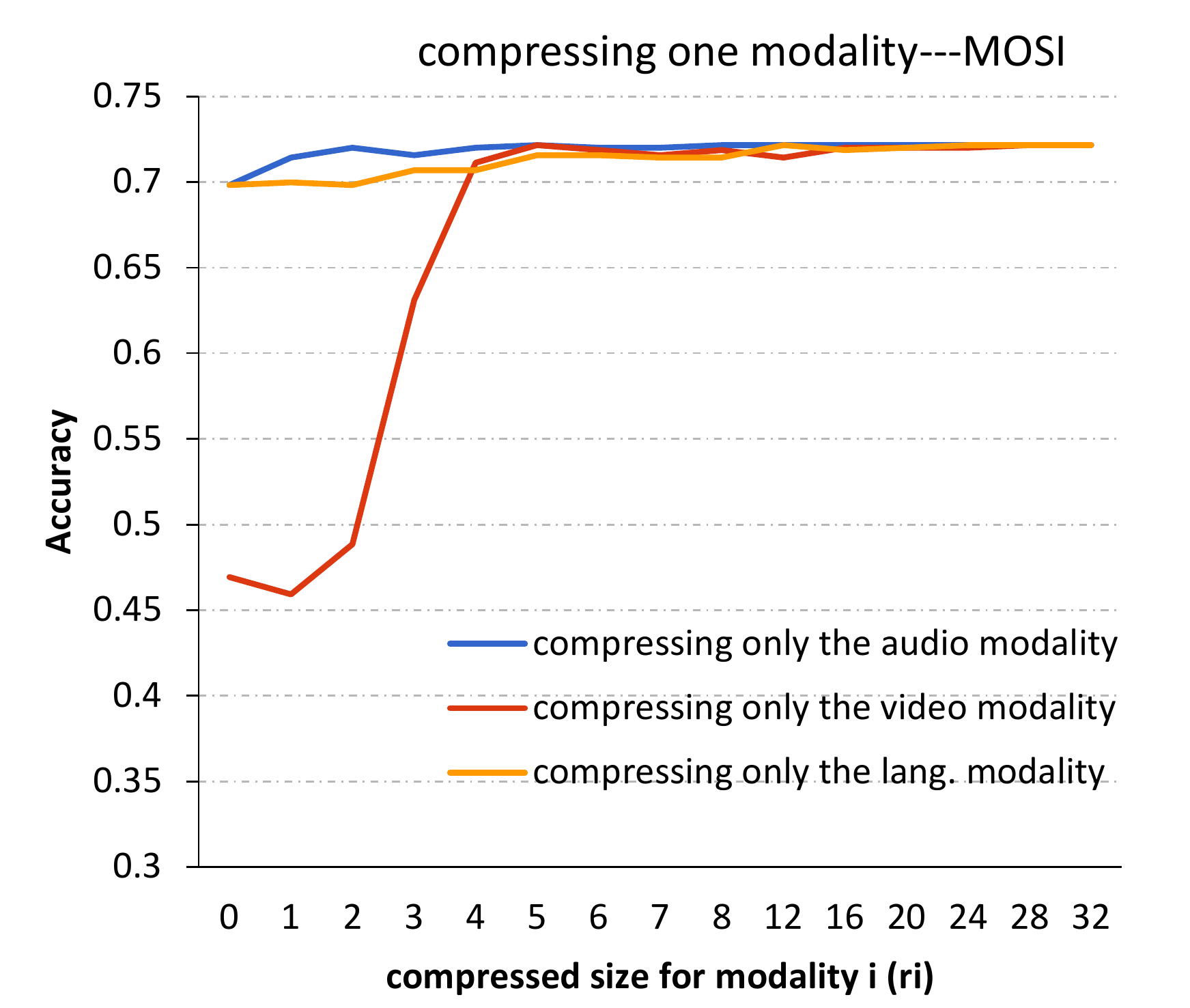}

     \caption{\label{fig:mosi}CMU-MOSI sentiment analysis dataset: Effect of different compression rates on accuracy for single modalities.} 
% \end{figure}
% \begin{figure}[!htb]
      \centering
      \hspace*{-.75cm} 
      \includegraphics[trim=2cm 3cm 10cm 2cm,clip, scale=0.4]{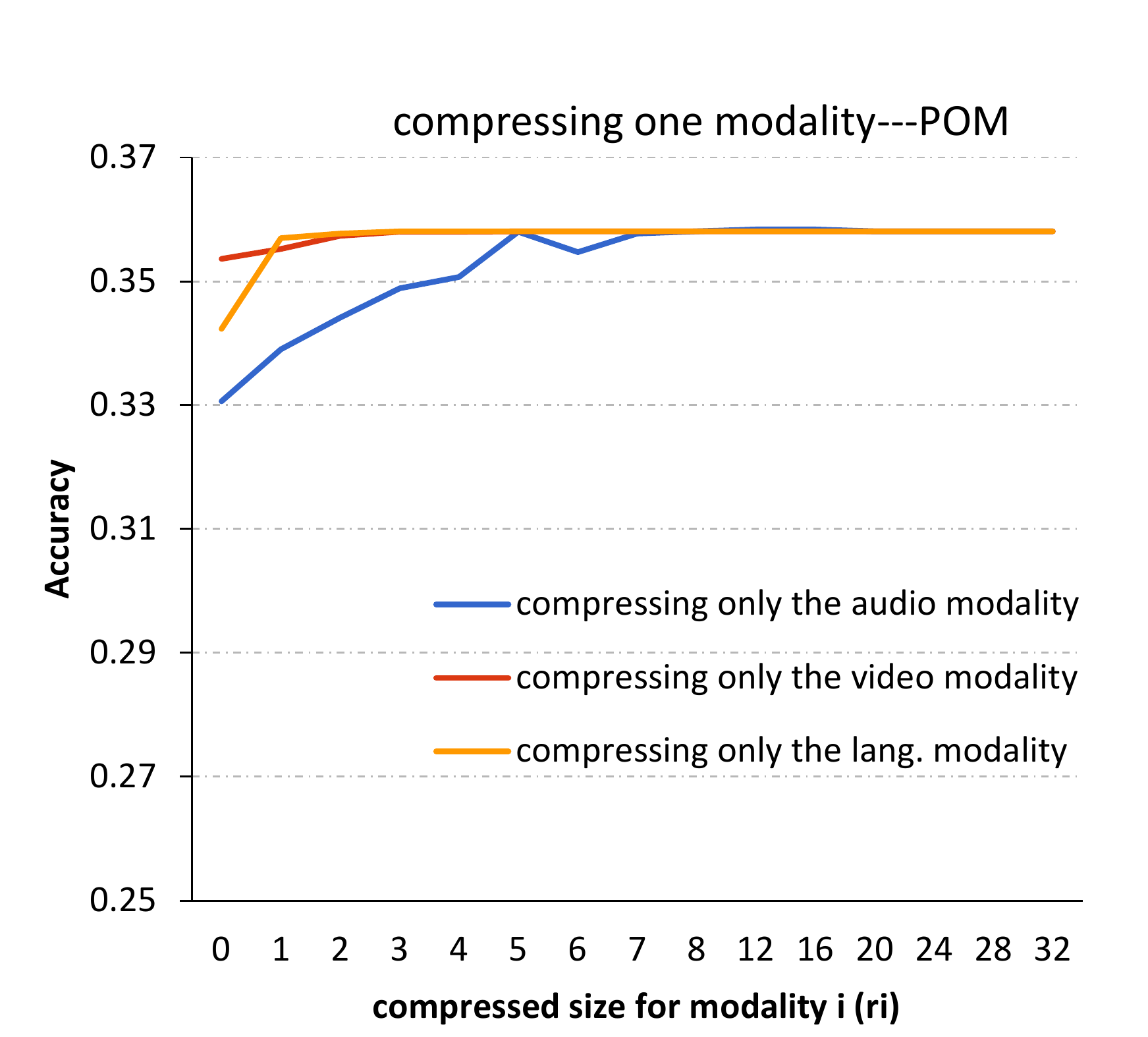}

\caption{\label{fig:pom}POM personality recognition dataset: Effect of different compression rates on accuracy for single modalities.}
\end{figure}

\begin{figure*}[h]
\centering
\begin{subfigure}[b]{0.49\textwidth}
\centering
% \hspace*{-.75cm}
\includegraphics[trim=2.5cm 10.2cm 7.5cm 11cm,clip, scale=0.64]{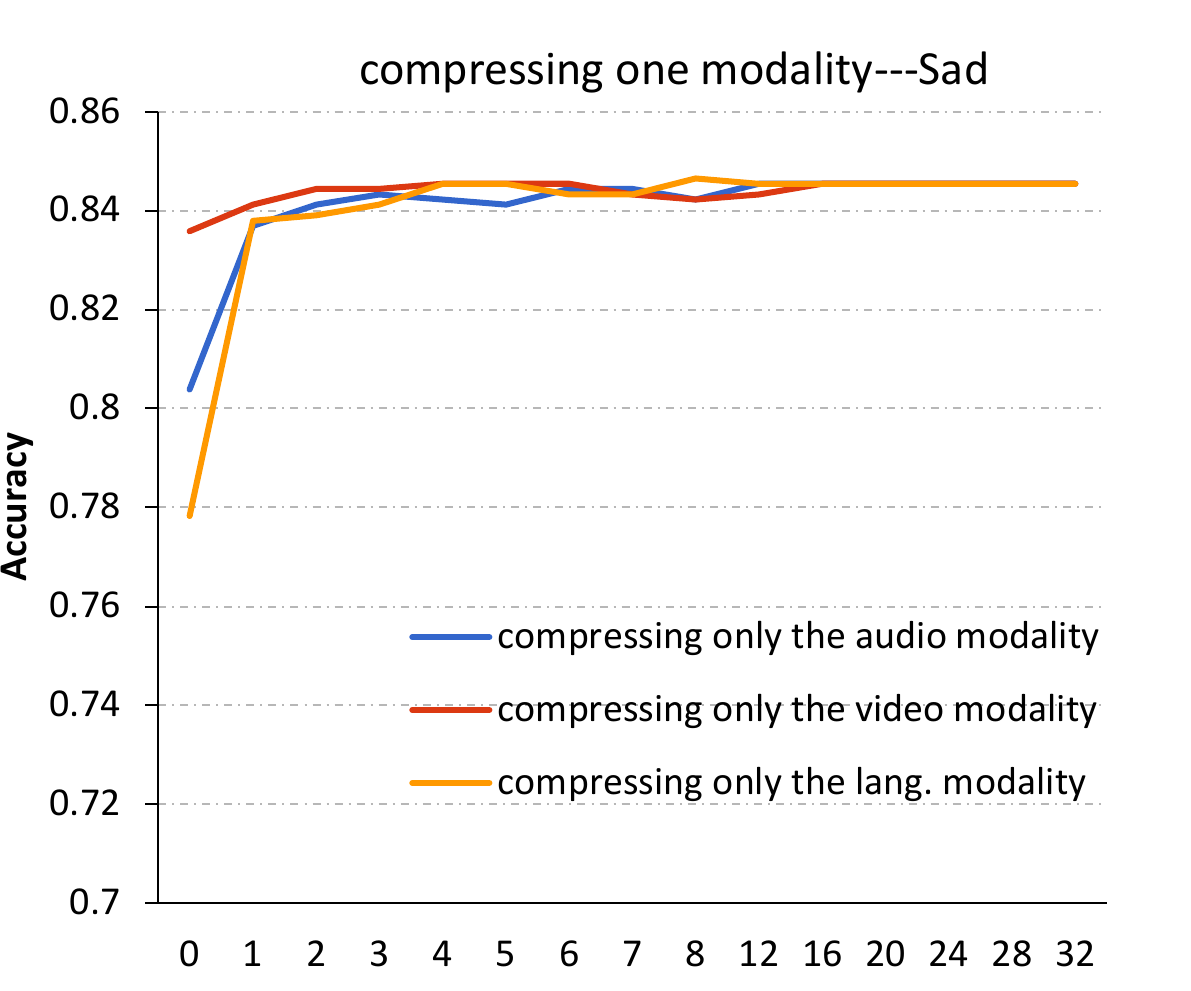}
\end{subfigure}
\hfill %%
\begin{subfigure}[t]{0.49\textwidth}
\centering
% \hspace*{-.75cm}
\includegraphics[trim=2.7cm 3.8cm 9.8cm 3cm,clip, scale=0.455]{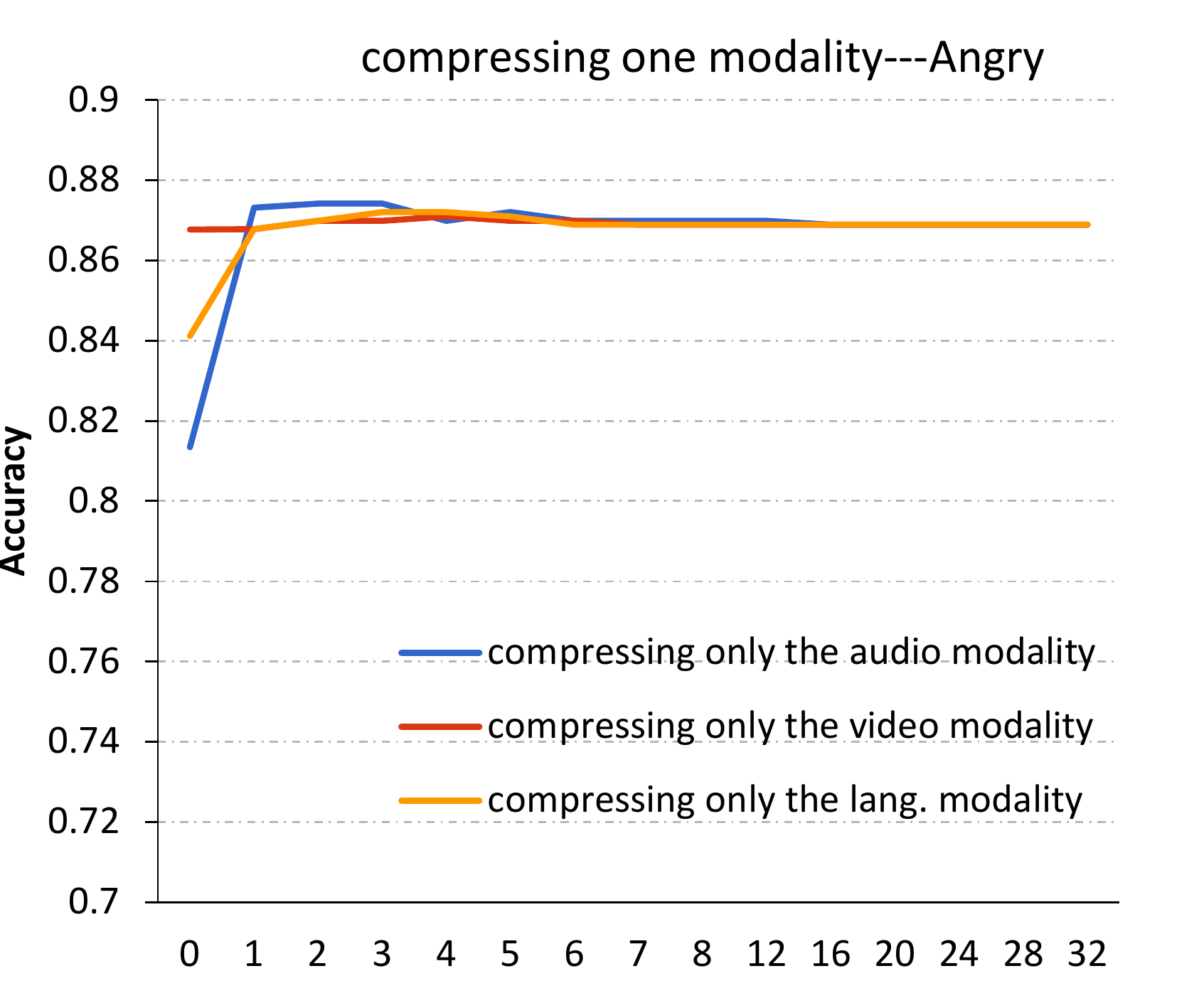}
\end{subfigure}

\begin{subfigure}[t]{0.49\textwidth}
\centering
% \hspace*{-.75cm}
\includegraphics[trim=2.5cm 9.5cm 7.5cm 11cm,clip, scale=0.64]{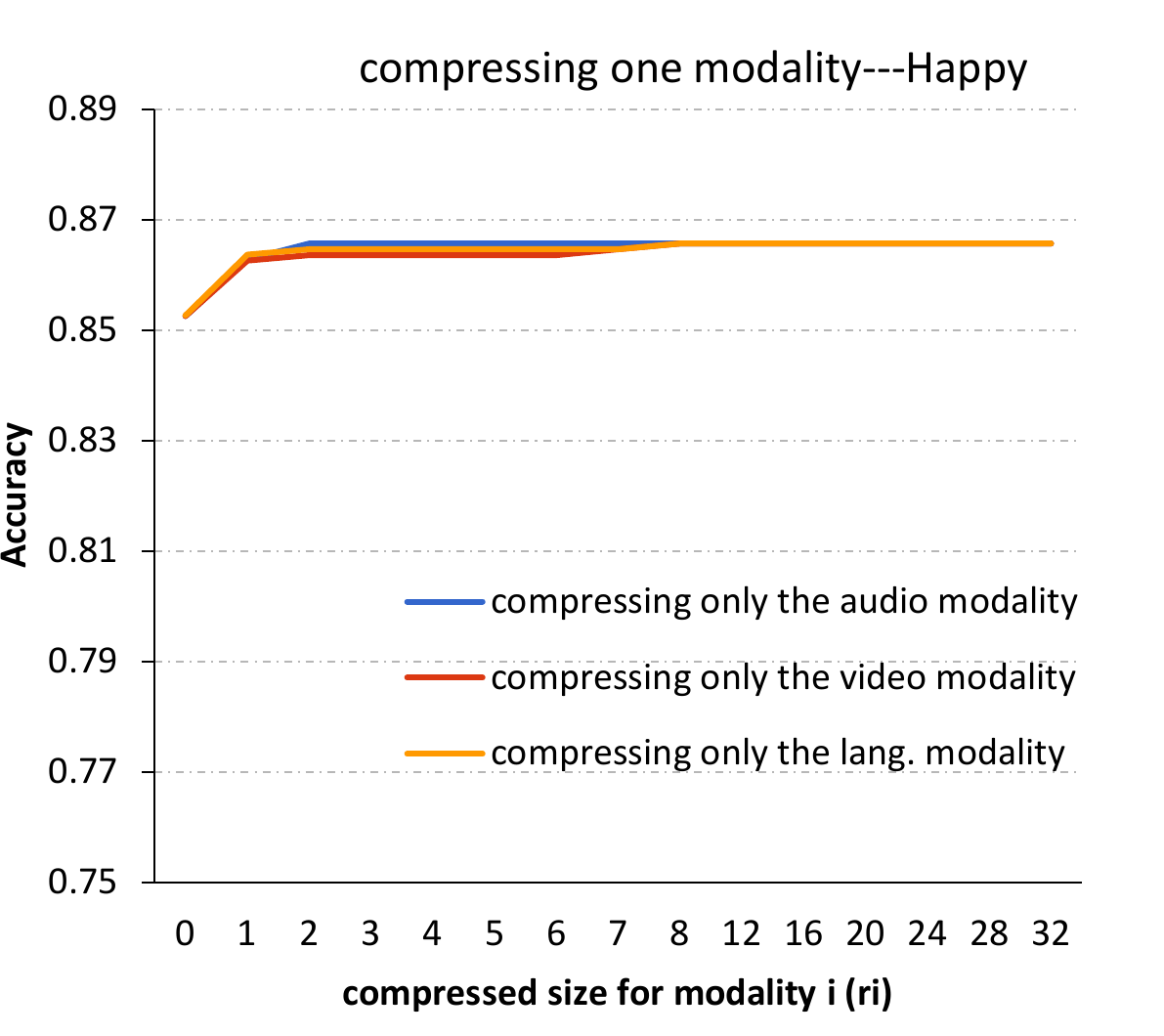}
\end{subfigure}
\hfill %%
\begin{subfigure}[t]{0.49\textwidth}
\centering
% \hspace*{-.75cm}
\includegraphics[trim=2.7cm 3cm 9.8cm 3cm,clip, scale=0.455]{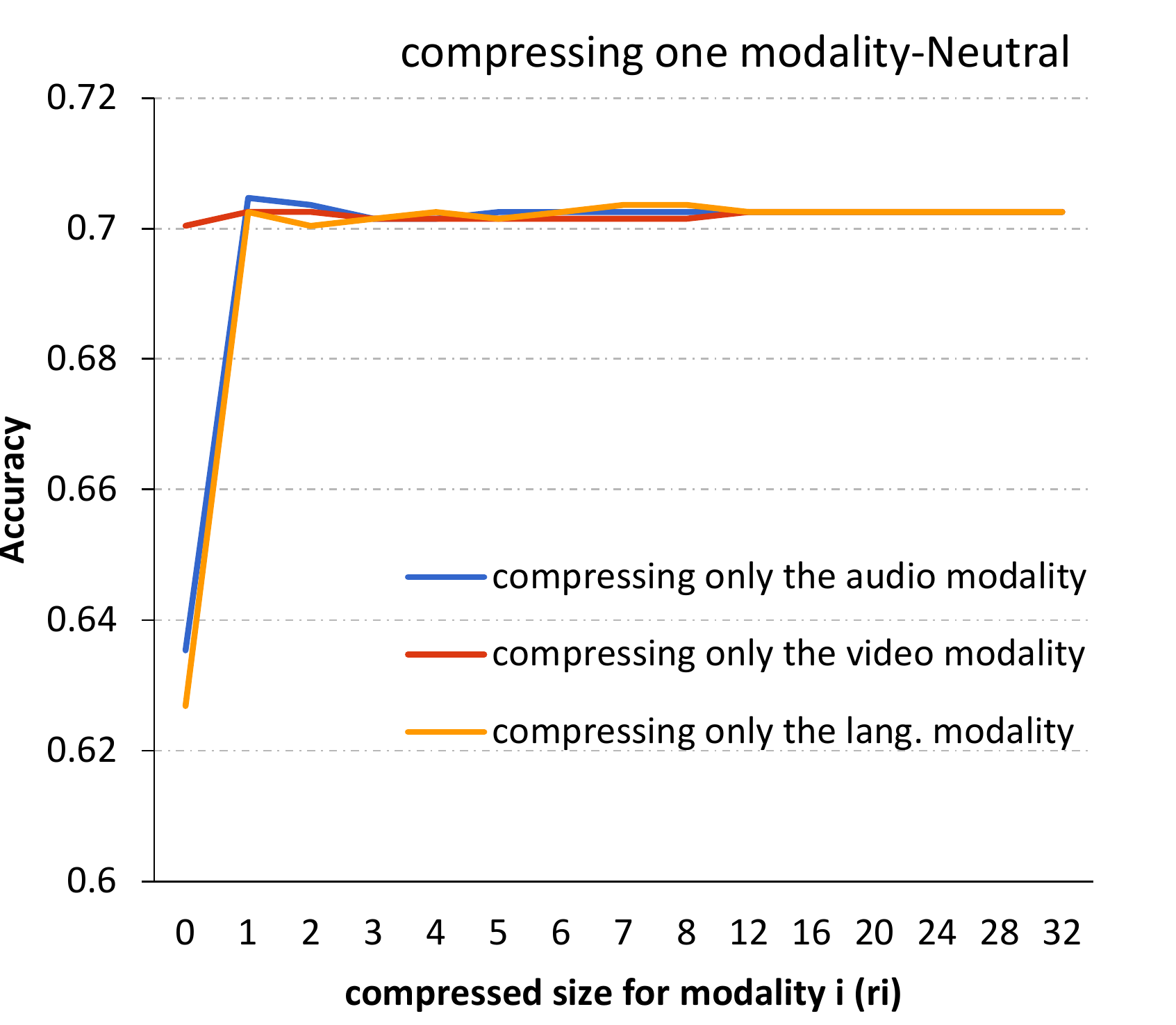}
\end{subfigure}
\caption{\label{fig:iemocap}IEMOCAP Emotion Recognition Dataset: Effect of different compression rates on accuracy for single modalities.}
\end{figure*}

The results can be seen in Fig. \ref{fig:mosi},  \ref{fig:pom} and \ref{fig:iemocap}. 
The horizontal axis is the compressed size and the vertical axis shows the accuracy for each modality. 
Note that due to the padding of each $D_i$ with 1, we have used $r_i+1$ as the new embedding size. 
% The ratio of the compressed size over the original size for the $i_{th}$ modality the compression rate can be calculated by $f_i=1-\frac{\text{Compressed size}}{\text{Original size}}=1-\frac{r_i}{d_i}$.

The first point that could be perceived clearly from the different modality diagrams is that each of the modalities changes in a different way when getting compressed, which means they each have a different amount of information that can not be compensated by the non-compressed modalities. 
In other words, a high accuracy when a modality is  
highly compressed means that there is a lot of redundant information in this modality --- 
the information loss resulting from factorization could be compensated by the other modalities so performance was not reduced. 

Fig. \ref{fig:mosi} shows results for the CMU-MOSI sentiment analysis dataset. For this dataset, a notable decrease in accuracy can be seen by compressing the video modality, while the audio and text modalities are not notably sensitive to compression. 
This shows that for sentiment analysis based on CMU-MOSI dataset, the information in Video modality cannot be compensated by other modalities, however most information in the audio and language modalities is covered in video modality. In other words, the video contains essential information for this task whereas information from audio and language can be recovered from video.  

Fig. \ref{fig:pom} shows the average accuracy over 16 personality types for the POM personality trait recognition dataset. For this dataset also, each of the modalities has a different behavior for different compression rates. 
We can see that the audio modality includes more non-redundant information for personality recognition as accuracy is highly affected by audio compression. 
In addition, there is a notable accuracy reduction when the language modality is highly compressed, which shows a small amount of non-redundant information for this task. 
Note that the POM data does not contain sufficient information for an effective analysis of the 16 personality sub types individually. 
% The graphs in Fig. \ref{fig:pom} show the average behavior across  personality types, however it also be interesting to perform an investigation for each personality type individually. 

Fig.~\ref{fig:iemocap} shows the results for the IEMOCAP emotion recognition dataset for each of the four emotional categories: happy, angry, sad, and neutral. 
Looking at the sad category, we see notable accuracy reduction for small sizes (high compression) for all the modalities, showing that each contains at least some non-redundant information. However, high compression of audio and especially language modalities results in strong accuracy reduction whereas video compression results in relatively minor reduction. It can be concluded that for this emotion, the language modality has the most non-redundant information and the video modality very little --- it's information can be compensated by the other two modalities. 
Moving on to the angry emotion, small sizes (high compression) result in accuracy reduction for audio and language modalities, showing that they contain some non-redundant information, with the audio modality containing more. Again the information in video can be almost completely compensated by the other two modalities. 

By comparing the highest accuracy values for various emotion categories, it is observed that neutral is hard to predict in comparison to the other categories. Again, the audio and Language modalities both include non-redundant information leading to a severe accuracy reduction with high compression of these modalities, with video containing almost no information not compensated by audio and language.  

The happy category is the easiest to predict emotion, and it slightly suffers for very small sizes of audio and video and language modalities, indicating a small amount of non-redundant information in all modalities.

\section{Conclusion} \label{sec:conclusion and Future Works} \label{sec:conclusion}
We proposed a tensor fusion method for multimodal media analysis by obtaining an $M+1$-way tensor to consider the high-order relationships between $M$ input modalities and the output layer. Our modality-based factorization method removes the redundant information in this high-order dependency structure and leads to fewer parameters with minimal loss of information. 
In addition, a modality-based factorization approach helps to understand the relative quantities of non-redundant information in each modality 
through investigation sensitivity to  
% useful information between modalities optimal
modality-specific compression rates. 
As the proposed compression method leads to a less complicated model, it can be applied as a regularizer which avoiding overfitting.  

We have provided experimental results for combining acoustic, text, and visual modalities for three different tasks: sentiment analysis, personality trait recognition, and emotion recognition.
We have seen that the modality-based tensor compression approach improves the results in comparison to the simple concatenation method, the tensor fusion method and tensor fusion using the same factorization rank for all modalities, as proposed in the LMF method. 
In other words, the proposed method enjoys the same benefits as the tensor fusion method and avoids suffering from having a large number of parameters, which leads to a more complex model, needs many training samples and is more prone to overfitting. We have investigated the effect of the compression rate on single modalities while fixing the other modalities helping to understand the amount of useful non-redundant information in each modality. Moreover, we have evaluated our method by comparing the results with state-of-the-art methods, achieving a 1\% to 4\% improvement across multiple measures for the different tasks.  

In future work, we will investigate the relation between dataset size and compression rate by applying our method to larger datasets. 
This will help to understand the trade-off between the model size and available training data, allowing more efficient training and avoiding under- and overfitting. 

As the availability of data with more and more modalities increases, both finding a trade-off between cost and performance and effective and efficient utilization of available modalities will be vital. 
% To be specific, does adding more modalities result in new information? 
% If so, does the amount of performance improvement worth the resulting computational and memory cost? 
Exploring compression methods promises to help identify and remove highly redundant modalities.

\section*{Acknowledgments}
This work was partially funded by grants \#16214415 and \#16248016 of the Hong Kong Research Grants Council, ITS/319/16FP of Innovation Technology Commission, and RDC 1718050-0 of EMOS.AI.
Thanks to Dr. Ian D. Wood and Peyman Momeni for their helpful commnets.
\FloatBarrier

\bibliography{acl2019}
\bibliographystyle{acl_natbib}

\end{document}